%% file: main.tex
\documentclass[10pt,twocolumn,letterpaper]{article}

\usepackage{wacv}      
\input{preamble}

\definecolor{wacvblue}{rgb}{0.21,0.49,0.74}
\usepackage[pagebackref,breaklinks,colorlinks,allcolors=wacvblue]{hyperref}
\usepackage{makecell}
\usepackage{tabularray}

\def\wacvPaperID{1563} 
\def\confName{WACV}
\def\confYear{2026}

\title{PixelNav: Towards Model-based Vision-Only Navigation with Topological Graphs}

\author{Sergey Bakulin\\
Skolkovo Institute of Science and Technology, Sber Robotics Center\\
Moscow\\
{\tt\small sergey.bakulin@skoltech.ru}
\and
Timur Akhtyamov\\
Skolkovo Institute of Science and Technology\\
Moscow\\
{\tt\small timur.akhtyamov@skoltech.ru}
\and
Denis Fatykhov\\
Skolkovo Institute of Science and Technology\\
Moscow\\
{\tt\small denis.fatykhoph@skoltech.ru}
\and
German Devchich\\
Skolkovo Institute of Science and Technology\\
Moscow\\
{\tt\small german.devchich@skoltech.ru}
\and
Gonzalo Ferrer\\
Skolkovo Institute of Science and Technology\\
Moscow\\
{\tt\small g.ferrer@skoltech.ru}
}

\newcommand{\sgpix}{^I\mathbf{p}^{\text{sg}}_t}
\newcommand{\obs}{O_t}
\newcommand{\sgim}{V^{\text{sg}}_t}
\newcommand{\travmask}{T_t}
\newcommand{\robst}[1]{^R \mathbf{x}_{#1}}
\newcommand{\robctrl}[1]{\mathbf{u}_{#1}}
\newcommand{\robpos}[1][]{^R\mathbf{p}_{#1}}
\newcommand{\impoint}[1][]{^I\mathbf{p}_{#1}}
\newcommand{\hcam}{h^{\text{cam}}}

\begin{document}
\maketitle
\input{sec/0_abstract}    
\input{sec/1_intro}
\input{sec/2_related_works}
\input{sec/3_materials_and_methods}
\input{sec/4_experiments}
\input{sec/5_conclusions}
{
    \small
    \bibliographystyle{ieeenat_fullname}
    \bibliography{main}
}

\end{document}

%% file: preamble.tex
\usepackage{float}

%% file: sec/0_abstract.tex
\begin{abstract}
This work proposes a novel hybrid approach for vision-only navigation of mobile robots, which combines advances of both deep learning approaches and classical model-based planning algorithms. Today, purely data-driven end-to-end models are dominant solutions to this problem. Despite advantages such as flexibility and adaptability, the requirement of a large amount of training data and limited interpretability are the main bottlenecks for their practical applications. To address these limitations, we propose a hierarchical system that utilizes recent advances in model predictive control, traversability estimation, visual place recognition, and pose estimation, employing topological graphs as a representation of the target environment. Using such a combination, we provide a scalable system with a higher level of interpretability compared to end-to-end approaches. Extensive real-world experiments show the efficiency of the proposed method.

The code will be released upon acceptance of the paper.
\end{abstract}

%% file: sec/1_intro.tex
\section{Introduction}
Classical metric SLAM-based navigation systems have been the dominant solution for mobile robots' navigation for decades. By relying on high-quality pre-built maps and the fusion of various sensors such as LiDARs, cameras, IMU, and GNSS, it enables the usage of advanced state estimation techniques and reliable controllers. However, this results in a high cost for such systems. Moreover, living creatures such as humans or animals significantly outperform them in terms of adaptability and flexibility, achieving near-perfect exploration and navigation relying solely on visual input.

Those factors lead to the born of a research branch referred to as visual navigation, and its extreme case - vision-only navigation. The goal of the vision-only navigation is to build a navigation system that relies solely on visual input in single- and multi-camera settings. Modern state-of-the-art models \cite{shahvint, sridhar2024nomad, gode2024flownav} are trained in the end-to-end Imitation Learning (IL) paradigm with a large combination of multiple datasets \cite{shah2022gnm}. However, real-world autonomous systems must be more interpretable and certifiable. Despite progress in the area of interpretable DL models \cite{li2022interpretable} and safety certification of deep control policies \cite{everett2021certifiable, cosner2022end, cheng2023prescribed}, there are still no commonly accepted solutions.

We are aiming to address this limitation by proposing a hierarchical system where final actions are produced by a Model Predictive Control (MPC) policy. Following previous works, topological graphs \cite{shah2021rapid, shahvint, chiang2024mobility} are used as a replacement for a classical dense metric map of the environment. Our high-level planner is responsible for the current graph node localization using the Visual Place Recognition (VPR) technique \cite{anyloc2023}. Then, given the current visual observation and the image from the subgoal node, the target pixel is selected on the observation image using pose estimation and traversability segmentation map \cite{kim2024learning}. The low-level planner employs projection equations to build a cost function that drives the robot towards the target pixel and ensures staying in the traversable region. We refer to the proposed method as $\textit{PixelNav}$, reflecting the usage of the pixel space for planning.

PixelNav achieves performance comparable to the state-of-the-art end-to-end navigation models and shows robustness to the novel obstacles missing in the original topological graph. The modular architecture enables continuous improvement of the method by identifying the sources of the issues and upgrading the corresponding components. Moreover, application of the model-based controller potentially enables analysis and certification techniques from traditional control theory.

\begin{figure*}[htbp]
    \centering
    \includegraphics[width=0.8\textwidth]{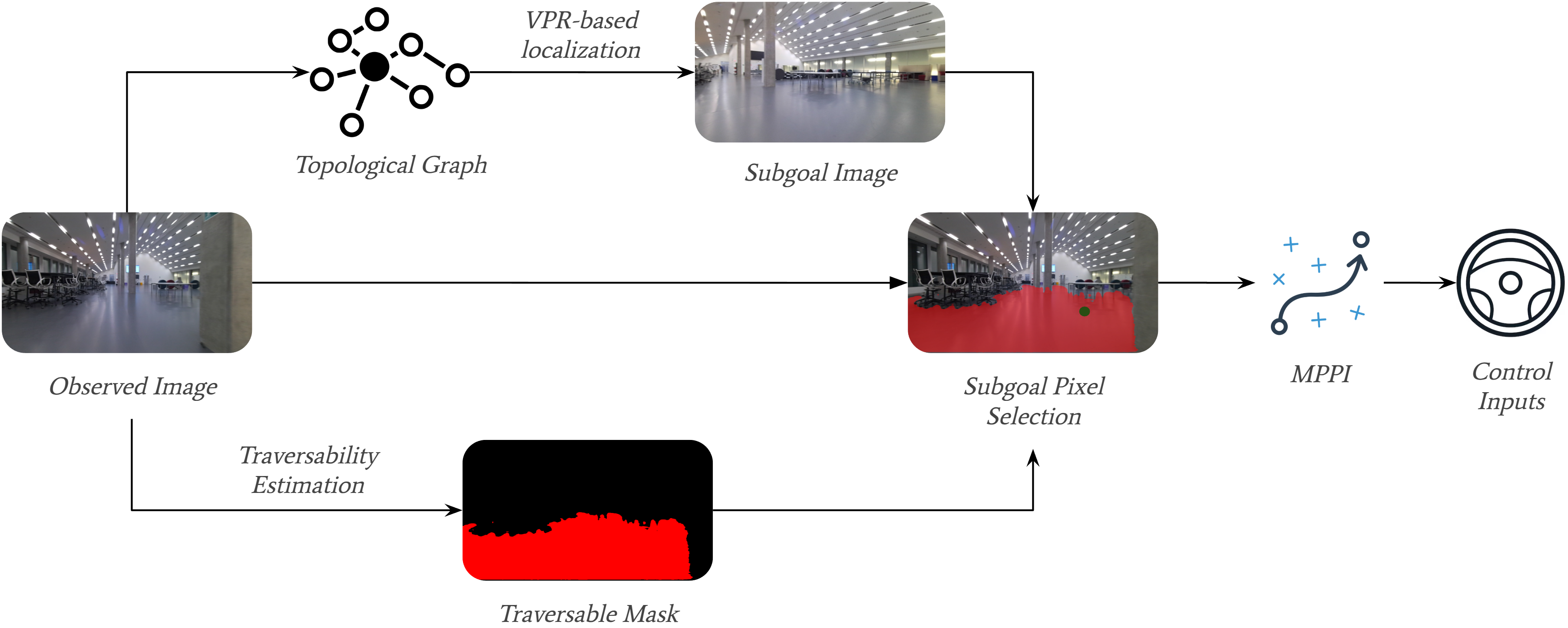}
    \caption{Overview of the method}
    \label{fig:visual_abstract}
\end{figure*}

%% file: sec/2_related_works.tex
\section{Related Works}
\textbf{Vision-Only Navigation.} Vision-Only Navigation is a promising direction that has been studied for several decades. Early works tried to solve this problem using classical computer vision concepts \cite{mccarthy2004performance, ohnishi2008visual}. Later, with the growing popularity of Deep Learning (DL), Reinforcement Learning (RL) approaches became the dominant paradigm \cite{kulhanek2019vision, zhu2017target, zeng2020survey}. However, RL-based methods are known to struggle with the sim-to-real problem \cite{dimitropoulos2022brief} that is a main bottleneck for real-world applications. With the growing amount of real robot \cite{shah2021rapid, karnan2022socially, nguyen2023toward, hirose2023sacson} and web-based \cite{liu2025citywalker, hirose2025lelan} datasets, the common paradigm was shifted towards the Imitation Learning (IL) in various robotics tasks, including navigation. Transformer-based models employed in a regression \cite{shahvint} or generative \cite{sridhar2024nomad, gode2024flownav} paradigms are the state-of-the-art real-world vision-only navigation models. In this work, we consider them as baselines. 

Alternative approaches tried to employ NeRFs \cite{adamkiewicz2022vision}, Gaussian splitting \cite{chen2025splat}, and video-based world models \cite{bar2025navigation}. The closest work to us is \cite{phan2025visionbasedperceptionautonomousvehicles}; however, it is focused on autonomous driving in road scenarios and uses depth estimation and object detection, while we target a larger set of scenarios and employ a more general traversability segmentation approach.

\textbf{Traversability estimation and traversability-aware navigation.} The concept of traversability estimation \cite{cai2022risk, cai2023probabilistic, cai2024evora, gasparino2022wayfast, gasparino2024wayfaster} was actively exploited in off-road navigation scenarios. The idea is to assign semantic classes and/or cost values to the regions of camera and/or LiDAR observations. Such maps are then used for planning, where a sampling-based version of MPC - Model Predictive Path Integral (MPPI) \cite{williams2017information} is employed thanks to its flexibility in supporting arbitrary forms of cost functions and nonlinear dynamics. Other methods also involve training RL policies with semantic inputs for general-purpose navigation \cite{roth2024viplanner}. A significant breakthrough in the traversability estimation field came with the release of the Segment Anything (SAM) model \cite{kirillov2023segment} which enabled self-supervised methods for traversability segmentation for the off-road \cite{jung2024v} and city/indoor scenarios \cite{kim2024learning} that do not require manual labeling. Standard segmentation models can then be trained with such datasets in a standard supervised paradigm.

\textbf{Topological graphs for navigation}. A topological graph is the sparse representation of the environment, which can be considered as an alternative to the classical dense maps. This graph includes a set of nodes - images of the different locations in the environment - and edges that define connectivity between them. Initially introduced for the virtual environments \cite{savinov2018semi}, this approach found natural applications in the field of visual navigation \cite{shah2021rapid, shah2021ving, shah2022viking, shahvint, chiang2024mobility}. Connectivity between the scenes is defined via approximate temporal \cite{shah2021rapid, shahvint} or spatial \cite{chiang2024mobility} distance measurement or heuristic. Localization, e.g., finding the closest scene to the currently observed scene, is performed using distance estimation \cite{shahvint} or via the VPR techniques \cite{chiang2024mobility}.

%% file: sec/3_materials_and_methods.tex
\section{Materials and Methods}

In this section, we provide a general overview of the system, along with details on its components - high-level and low-level planners.

\subsection{General Overview}

A general overview is provided in Fig. \ref{fig:visual_abstract}. The goal of the system is to provide the robot-specific control input $\mathbf{u}_t$ that will safely drive the robot towards the destination given the image $O_t$ observed by the camera for each discrete decision-making time step $t$. We utilize a topological graph \cite{shah2021rapid, shah2022gnm, chiang2024mobility} $\mathcal{G} = \{ \mathcal{V}, \mathcal{E} \}$ as a representation of the target environment, where $\mathcal{V} = \{ V\left[i \right] \text{ } | \text{ } i = 1, \dots, N_{\mathcal{V}} \}$ is a set of image nodes and $\mathcal{E} = \{ E\left[j \right] \text{ } | \text{ } j = 1, \dots, N_{\mathcal{E}} \}$ is a set of edges which define connectivity between scenes. This connectivity can be defined as an estimate of scaled / unscaled temporal \cite{shahvint} or spatial \cite{chiang2024mobility} distance estimation; the latter is used in this work. Destination is defined as a goal image node $V^{\text{goal}} \in \mathcal{V}$ selected by the user or another system.

At each \textit{relocalization} step, graph localization is performed by finding temporally \cite{shahvint}, semantically \cite{chiang2024mobility} or spatially closest node $V^{\text{loc}}_t \in \mathcal{V}$ to the current observation $O_t$. Next, the standard graph pathfinding algorithm like Dijkstra's / A$^*$ finds a node sequence $\mathcal{P}_t = \{ V^{\text{path}}_t \left[ l \right] \text{ } | \text{ } l = 1, \dots, |\mathcal{P}_t|; V^{\text{path}}_t \left[ l \right] \in \mathcal{V}\}$ that leads from $V^{\text{loc}}_t$ to $V^{\text{goal}}$. A node with some offset index $l_{\text{sg}}$ is selected as a \textit{subgoal} image $V_t^{\text{sg}} := V^{\text{path}}_t \left[ l_{\text{sg}} \right]$.

Given an observation image $O_t$ and a subgoal image $V^{\text{sg}}_t$, the \textit{subgoal pixel} which belongs to the \textit{traversable mask} $T_t$ is selected as a target for the low-level planner. The traversable mask $T_t$ is defined as a set of pixels of $O_t$ that belong to the projection of the safe and obstacle-free regions of the environment's surface. This mask is obtained with a deep traversability estimation model $\tau$: $T_t = \tau \left( O_t \right)$. The subgoal pixel $^{I}\mathbf{p}^{\text{sg}}_t$, where $I$ stands for the image plane coordinate system, is selected using heuristics described in the sections below.

Finally, the MPC-based low-level planner (controller) uses the previously obtained traversable mask $T_t$ and subgoal pixel $^{I}\mathbf{p}^{\text{sg}}_t$ to produce control input $\mathbf{u}_t$. In our work, a unicycle robot model with linear and angular velocities as control inputs is considered; however, the proposed method can be adapted to other models.

One should note that subgoal image $V_t^{\text{sg}}$ is performed at a lower rate compared to the other parts of the pipeline, e.g., the same $V_t^{\text{sg}}$ is re-used for multiple time steps $t$.

The next subsections provide details on all parts of the proposed pipeline.

\subsection{Topological Graph Construction and Localization}

To build a topological graph, first an \textit{ordered set} of images $\mathcal{I} = \{ I\left[ m \right] \text{ } | \text{ } m = 1, \dots, |\mathcal{I} | \}$ is collected and passed through the SLAM/odometry pipeline to produce a set of 2D-positions $\mathcal{Z} = \{ ^{\hat{W}}\mathbf{z} \left[ n \right] \text{ } | \text{ } n = 0, \dots, |\mathcal{I}| \}$. To preserve the vision-only setting, the DPVO \cite{dpvo2022} visual odometry method is applied to estimate scale-free poses; $\hat{W}$ denotes this scale-free world frame. The graph $\mathcal{G}$ is constructed using $\mathcal{I}$ and $\mathcal{Z}$. 

The set of images $\mathcal{I}$ is directly converted to $\mathcal{V}$ with optional downsampling; the order is preserved, and $^{\hat{W}}\mathbf{z}\left[ i \right]$ is used to denote a pose for an image node $V\left[i\right]$. A \textit{relative direction} $\phi\left[i\right]$ is also calculated for each pose:
\begin{equation}
    \label{eq:rel_direction}
    \phi \left[i\right] = \text{arctan2} \left( \frac{^{\hat{W}}\mathbf{z}\left[ i+1 \right] - ^{\hat{W}}\mathbf{z}\left[ i \right]}{\| ^{\hat{W}}\mathbf{z}\left[ i+1 \right] - ^{\hat{W}}\mathbf{z}\left[ i \right] \|_2} \right).
\end{equation}
An edge $E$ between nodes $V\left[i\right]$ and $V\left[i'\right]$ is added if they satisfy the following criteria:
\begin{enumerate}
    \item Euclidean criterion:
    \begin{equation}
        \label{eq:euclidean_criterion}
        \| ^{\hat{W}}\mathbf{z}\left[ i \right] - ^{\hat{W}}\mathbf{z}\left[ i' \right] \|_2 < \rho \mu,
    \end{equation}
    where $\mu$ is the mean of non-zero distances between two consecutive poses $^{\hat{W}}\mathbf{z}\left[ i \right]$ and $^{\hat{W}}\mathbf{z}\left[ i + 1 \right]$, and $\rho$ is a tunable coefficient;
    \item Angular criterion:
    \begin{equation}
        \label{eq:angular_criterion}
        | \phi \left[ i \right] - \phi \left[ i' \right] | < \phi_{\text{max}},
    \end{equation}
    where $\phi_{\text{max}}$ is a tunable threshold.
\end{enumerate}
These criteria help to ensure transition feasibility within a unicycle robot model.

Inspired by \cite{chiang2024mobility}, we use a VPR technique for localization instead of using a learned heuristic $\cite{shahvint}$. We employ the AnyLoc \cite{anyloc2023} approach based on DINOv2 \cite{oquab2024dinov2} features, particularly its configuration with Generalized Mean (GeM) pooling, since we found it more suitable for on-board deployment.

\subsection{Traversability Estimation}
\begin{figure}[htbp]
    \centering
    \begin{subfigure}[b]{0.9\linewidth}
        \centering
        \includegraphics[width=0.9\textwidth]{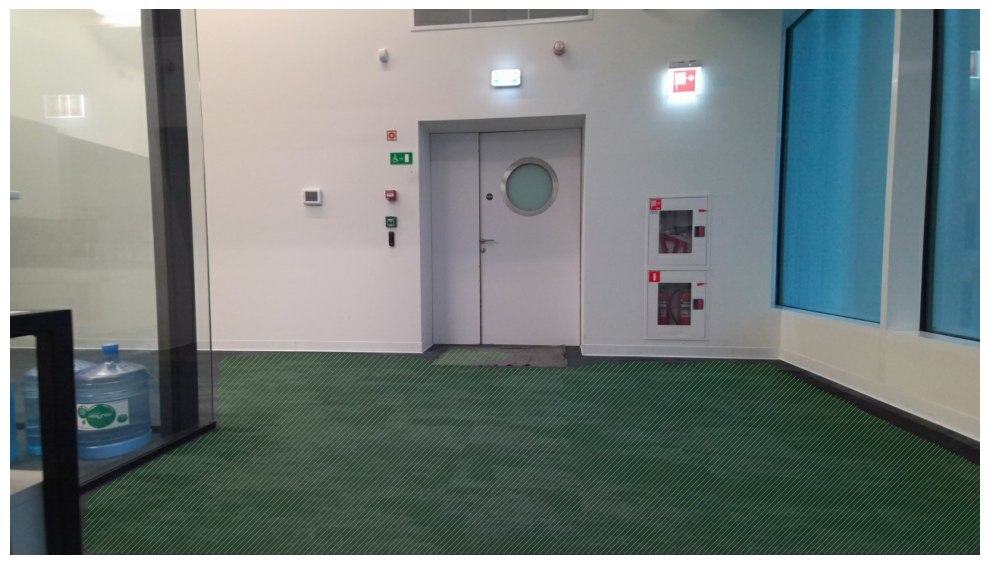}
    \end{subfigure}
    
    \vspace{0.5cm} 
    
    \begin{subfigure}[b]{0.9\linewidth}
        \centering
        \includegraphics[width=0.9\textwidth]{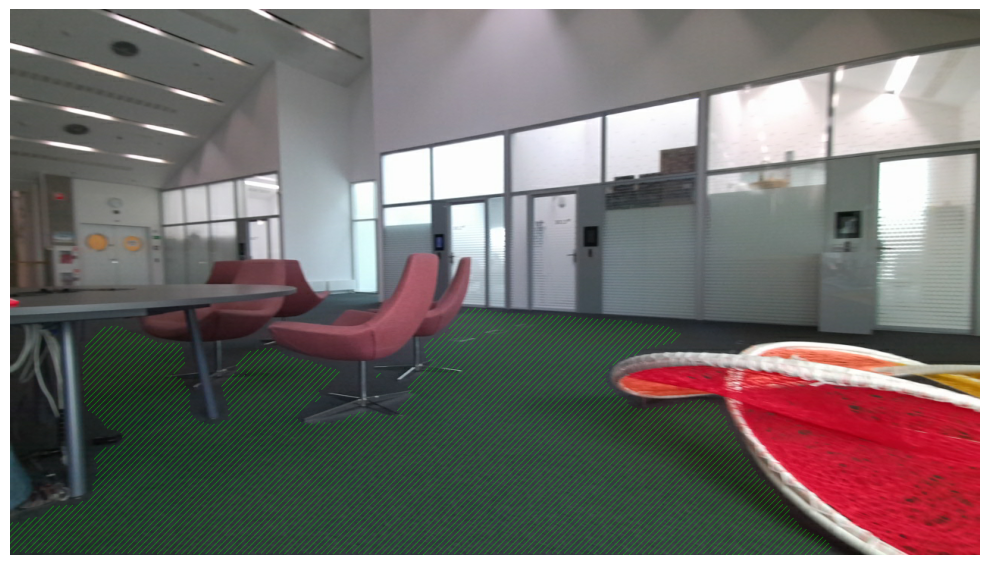}
    \end{subfigure}
    
    \caption{Example of the traversability prediction. The green mask defines the traversable region.}
    \label{fig:trav_good_examples}
\end{figure}

Traversability estimation is utilized for the subgoal pixel selection and the low-level controller to ensure movement only within the obstacle-free regions. Following \cite{kim2024learning, akhtyamov2025egowalk}, we employ a SegFormer-based binary segmentation model \cite{xie2021segformer} as a traversability segmentation model $\tau$. This model is trained on the subset of the EgoWalk traversability dataset \cite{akhtyamov2025egowalk}, which is an automatically labeled dataset by the SAM-based methodology \cite{kim2024learning}. The input of the model is the current image observation $O_t$, and the output is a binary mask $T_t$, where positively labeled pixels correspond to the traversable regions of the environment. An example of traversability estimation is shown in Fig. \ref{fig:trav_good_examples}.

\subsection{Subgoal Pixel Selection}
Given the current observation $\obs$ and the subgoal image $\sgim$, our goal is to select a target pixel $\sgpix$. For that, we build an algorithm based on the rigid body transformation between $\obs$ and $\sgim$ along with a traversable mask $\travmask$.

Assuming that $\obs$ and $\sgim$ are produced by the same camera with known intrinsic parameters, an essential matrix is calculated using RANSAC \cite{fischler1981random} for the set of keypoints detected by the SuperPoint model \cite{detone2018superpoint} and matched by the SuperGlue model \cite{sarlin2020superglue}. The rotation matrix is calculated based on the essential matrix, and the yaw component is extracted. This results in a relative rotation angle $\alpha_t$, which indicates a desired update of the robot's heading to align the observed scene with the subgoal scene.

\begin{figure}[H]
    \centering
    \includegraphics[width=0.47\textwidth]{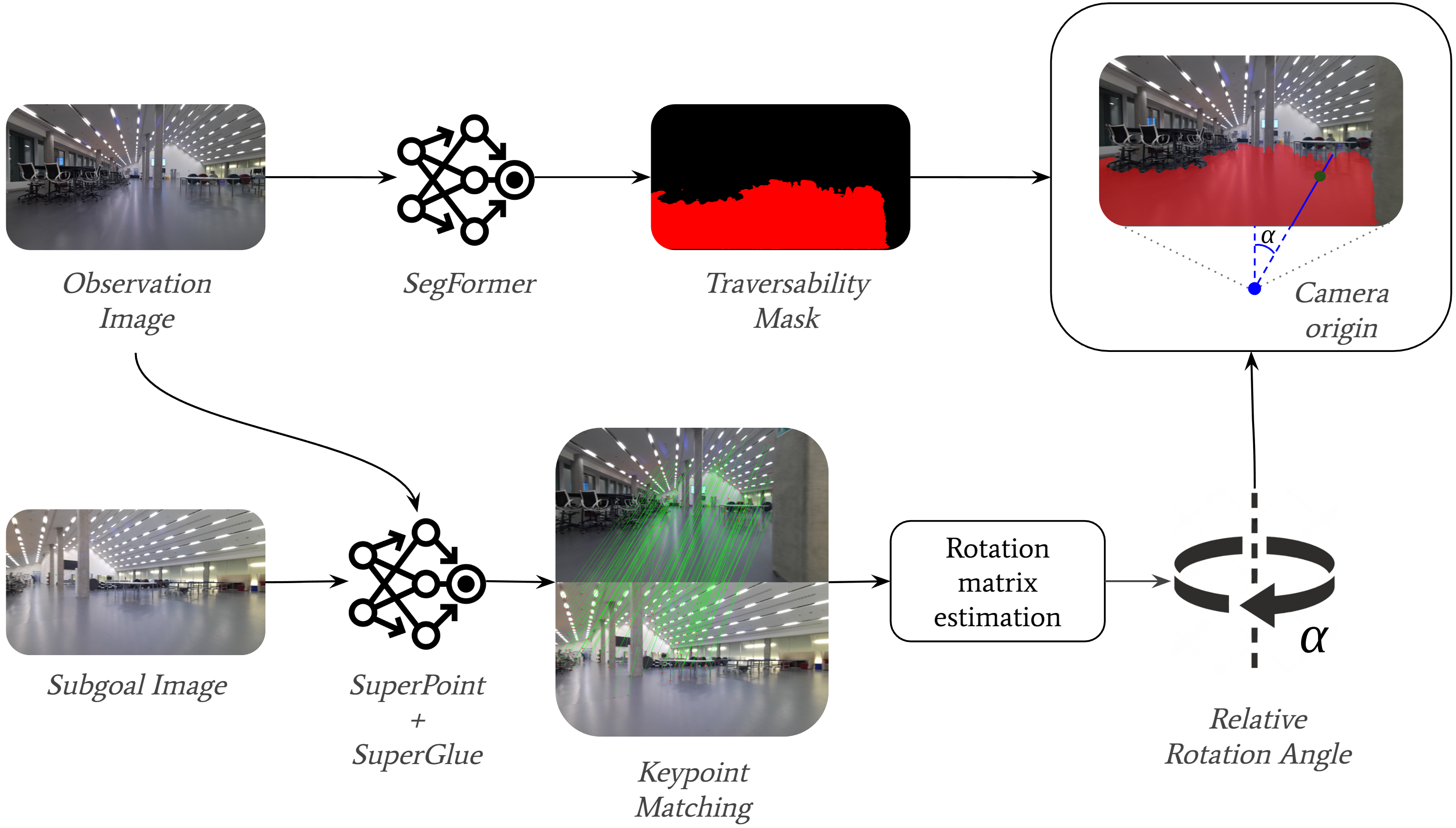}
    \caption{Subgoal pixel selection algorithm illustration. The selected pixel is marked in green.}
    \label{fig:sg_pixel_selection}
\end{figure}

Given the rotation angle, a ray is traced through the traversable mask at the angle $\alpha_t$, and its intersection with the upper border of this mask is found. We take a pixel that bounds the obtained segment within approximately 2/3 of the length to avoid ``dangerous'' points at the edge of the traversable masks. This process is illustrated in Fig. \ref{fig:sg_pixel_selection}. If the ray does not cross any traversable region, we take the closest pixel from the whole traversable mask to this ray.

This selected subgoal pixel $\sgpix$ is then fed as a target for the low-level planner for producing final control inputs.

\subsection{Low-level Planning}
Low-level planning leverages sampling-based MPPI policy \cite{williams2017information}. For developing the model and cost functions, several assumptions must be introduced:
\begin{enumerate}
    \item Camera is rigidly attached to the robot, its height $h^{\text{cam}}$ above the ground surface is fixed and known;
    \item Camera's intrinsic parameters are known;
    \item Camera's plane is orthogonal to the ground surface.
\end{enumerate}
While these assumptions may limit practical applications, intuitively, they are held for the majority of indoor and outdoor scenarios.

The state of the robot at the horizon step $k$ is defined as position and orientation in the robot's local frame $R$: $^R\mathbf{x}_k = \begin{bmatrix} ^Rx_k, ^Ry_k, ^R\theta_k \end{bmatrix}^\top$. Position part of the state is defined as $^R\mathbf{p}_k = \begin{bmatrix} ^R x_k, ^R y_k\end{bmatrix}^\top$. The unicycle kinematic model $F$ is employed inside MPPI:
\begin{equation}
    \label{eq:unicycle}
    \robst{k+1} = f(\robst{k}, \robctrl{k}) = \begin{bmatrix}
        ^Rx_k \\ ^Ry_k \\ ^R\theta_k
    \end{bmatrix} + \begin{bmatrix}
        v_k \cos (^R\theta_k ) \Delta t \\
        v_k \sin (^R\theta_k ) \Delta t \\
        w_k \Delta t
    \end{bmatrix},
\end{equation}
where $\robctrl{k} = \begin{bmatrix} v_k, w_k \end{bmatrix}^\top$ is a control input of linear and angular velocity; $\Delta t$ is a discrete planning time step. Note that for 2D configurations space, which is discussed here, we use Robot Operating System (ROS) compatible coordinate frame notation (X forward, Y left, Z up), while for the 3D case below OpenCV standard will be used (X-right, Y-down, Z-forward).

Assume that some point $\robpos = \begin{bmatrix} ^Rx & ^Ry \end{bmatrix}^\top$ is located on the ground surface. From the 3D perspective, it means that the same point in 3D frame will have coordinates (considering frame transforms described above):
\begin{equation}
    \label{eq:known_3d}
    ^C\mathbf{p}^{\text{3D}} = \begin{bmatrix}
        -^Ry \\
        h^{\text{cam}} \\
        ^Rx
    \end{bmatrix},
\end{equation}
where $C$ defines the camera frame.
Thus, with a known camera matrix $K$, the projection $^I\mathbf{p} = \begin{bmatrix} ^I u & ^I v\end{bmatrix}^{\top}$ can be calculated in a standard way:
\begin{equation}
    \label{eq:projection}
    \begin{bmatrix}
        ^I \mathbf{p} \\
        1
    \end{bmatrix} = K \begin{bmatrix}
        -^Ry / ^Rx \\
        h^{\text{cam}} / ^Rx \\
        1
    \end{bmatrix}.
\end{equation}
This operation, together with extracting $\impoint$ is denoted by a function $P$:
\begin{equation}
    \label{eq:projection_fn}
    \impoint = P(\robpos)
\end{equation}

Vice versa, if it is known that the pixel $^I \mathbf{p}$ belongs to the ground surface, the inverse problem becomes well-determined, and the corresponding point $\robpos$ can be found. For simplicity, we denote this operation as $P^{-1}$:
\begin{equation}
    \label{eq:}
    \robpos = P^{-1} (\impoint) = \begin{bmatrix}
        \frac{f_y \hcam}{^Iv - c_y} \\
        -\frac{(^Iu - c_x) f_h \hcam}{f_x (^Iv - c_y)}
    \end{bmatrix},
\end{equation}
where $f_x$, $f_y$, $c_x$ and $c_y$ are the components of the camera matrix $K$. This operation is also known as an Inverse Perspective Mapping (IPM).

We utilize those concepts to construct the cost function for MPPI, which is a weighted sum of two components, subgoal-reaching cost and collision avoidance cost. The subgoal-reaching cost is responsible for driving the robot towards the subgoal scene by minimizing the distance between the subgoal pixel and the future robot positions in the pixel space of $O_t$:
\begin{equation}
    \label{eq:goal_reaching_cost}
    q^{\text{sg}} (\robst{k}) = \| P(\robst{k}) - \sgpix \|_2.
\end{equation}
Intuitively, this cost acts as a heuristic that drives the robot towards the desired scene by following a point whose azimuth is correlated with the desired pose transformation.

\begin{figure}[htbp]
    \centering
    \begin{subfigure}[b]{0.9\linewidth}
        \centering
        \includegraphics[width=0.9\textwidth]{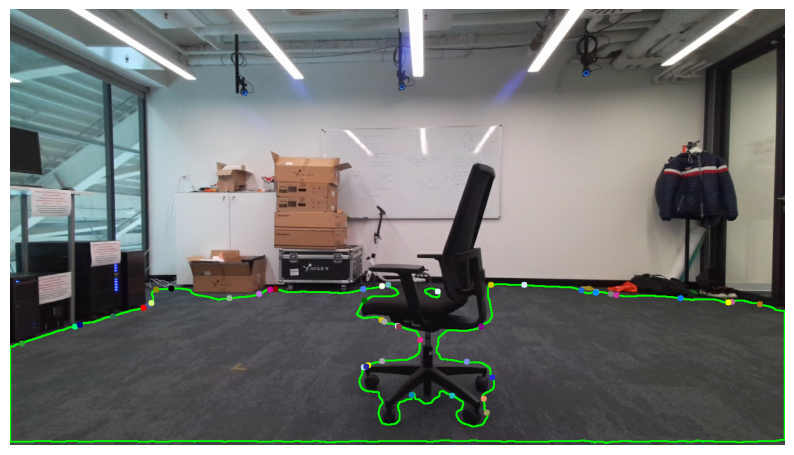}
        \caption{Contour points in the image space}
        \label{fig:contours_example}
    \end{subfigure}
    
    \vspace{0.5cm} 
    
    \begin{subfigure}[b]{0.9\linewidth}
        \centering
        \includegraphics[width=0.9\textwidth]{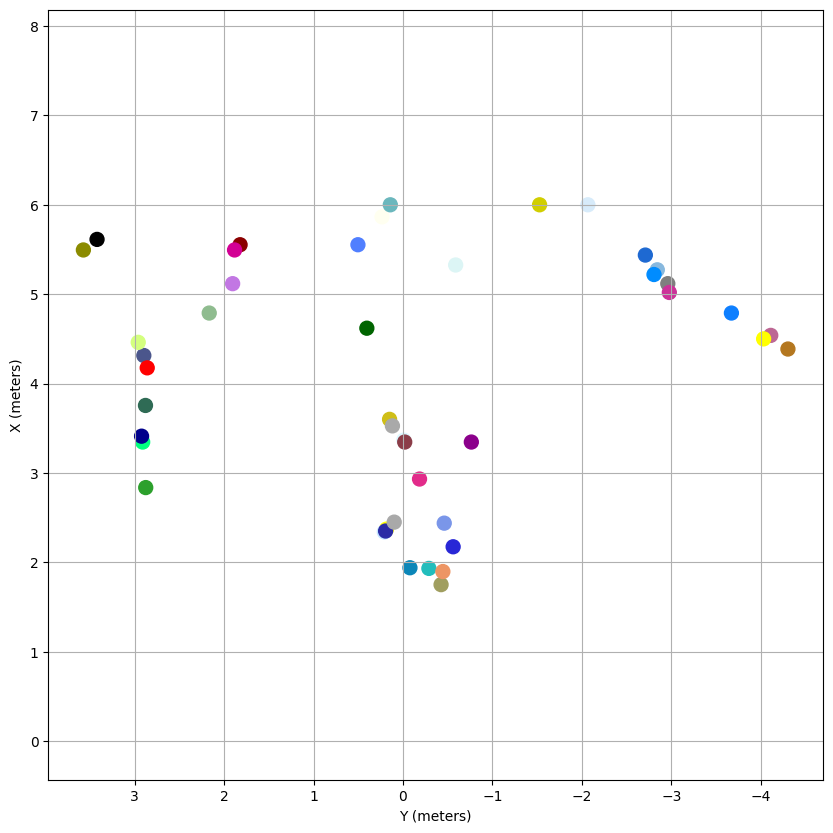}
        \caption{IPM-based backprojection of the contour points}
        \label{fig:ipm_example}
    \end{subfigure}
    
    \caption{Example of the contour points sampling and IPM}
    \label{fig:contours_ipm_example}
\end{figure}

One more concept needs to be introduced to define the collision avoidance cost. For the current traversable mask $T_t$, a set of contours is extracted using the standard Suzuki algorithm \cite{suzuki1985topological}. For each contour, a set of random points is sampled. After combining those sets, we obtain a set of \textit{obstacle} points $\mathcal{C}^{\text{obst}}_t = \{ ^I \mathbf{p}^{\text{obst}}_t \left[ m \right] \ \text{ } | \text{ } m = 1, \dots, N^{\text{obst}}_t \}$. Collision avoidance cost is defined using IPM:
\begin{multline}
    \label{eq:collision_cost}
     q^{\text{obst}} (\robst{k}) = \\ = \sum_{m=1}^{N^{\text{obst}}_t} \mathbf{1} \left( \| ^R \mathbf{p}_{k} - P^{-1}(^I \mathbf{p}^{\text{obst}}_t \left[ m \right]) \|_2 \right.
     \left. < r^{\text{safe}} \right)
\end{multline}
where $\mathbf{1}$ is the indicator function, $r^{\text{safe}}$ is the collision threshold distance, which is defined by the approximate radius of the robot and the desired safety level. An example of the described sampling is shown in Fig. \ref{fig:contours_ipm_example}.

The final cost is the weighted sum of these costs plus the penalty on control inputs:
\begin{multline}
    \label{eq:full_cost}
    q(\robst{k}, \robctrl{k}) = w_{\text{obst}} q^{\text{obst}} (\robst{k}) + w_{\text{sg}}q^{\text{sg}} (\robst{k}) + \\ + \robctrl{k}^{\top} Q_{\text{ctrl}} \robctrl{k},
\end{multline}
where $w_{\text{obst}}$, $w_{\text{sg}}$ are the weight scalars, $Q_{\text{ctrl}}$ is the weight matrix.
Note that here we mix cost components defined in both the image plane and the world coordinate frame. Based on our observations, this combination gave the best results during real-world debugging and tuning.

%% file: sec/4_experiments.tex
\section{Experiments}
In this section, we present the evaluation procedure of PixelNav and several baseline methods, as well as the experimental results and their analysis. Our goal is to understand the capabilities and limitations of the method in a real-world navigation task. The readers are encouraged to get familiar with the supplementary videos\footnote[1]{\url{https://www.youtube.com/playlist?list=PLzwD1_-1fQT3MLLFd4sitaSZMKOLdixgI}} to better understand the methodology and perform qualitative analysis.

\subsection{Implementation Details}
The proposed method is implemented using C++ and Python on top of the ROS 2 framework \cite{doi:10.1126/scirobotics.abm6074}. Graph localization, traversability estimation, subgoal pixel selection, and the MPPI controller are implemented in separate nodes to ensure efficient parallel execution. The traversability estimation model and the SuperPoint and SuperGlue models are deployed using the TensorRT framework to enhance their onboard computational performance. For the AnyLoc-based localization, we use GeM pooling instead of NetVLAD \cite{arandjelovic2016netvlad} due to the large size and computational complexity in terms of a near-realtime application. The main values of parameters are presented in Table \ref{tab:model_parameters}.

The mobile robot used in the experiments is built on top of the AgileX Tracer platform and equipped with \textit{Intel NUC11PHKI7C} (\textit{Nvidia 2060} 6 Gb GPU) computational module and \textit{Azure Kinect} camera (depth channel was not used). 

\begin{table}[htbp]
\centering
\caption{Deployment parameters}
\label{tab:model_parameters}
\begin{tabular}{cl}
\hline
\textbf{Parameter}                                   & \textbf{Value} \\ \hline
\makecell[l]{\text{Traversability binary}\\\text{segmentation threshold}} & $0.95$                               \\
$w_{\text{obst}}$                                                   & $10$                                 \\
$w_{\text{sg}}$                                                   & $10$                                 \\
$Q_{\text{ctrl}}$                                                    & $\text{diag}(1, 100)$                             \\
$r^{\text{safe}}$                                                    & $2$                                   \\
$\Delta t$                                                    & $0.2$                                    \\ \hline
\end{tabular}
\end{table}

\subsection{Methodology}
We enhance the commonly used evaluation procedures focused on the goal reaching and collision avoidance of navigation models \cite{shahvint, sridhar2024nomad, gode2024flownav} by introducing perturbation techniques and two classes of collision cases to ensure a deeper understanding of the method's behavior. 

\begin{figure}[htbp]
    \centering
    \begin{subfigure}[b]{0.9\linewidth}
        \centering
        \includegraphics[width=0.9\textwidth]{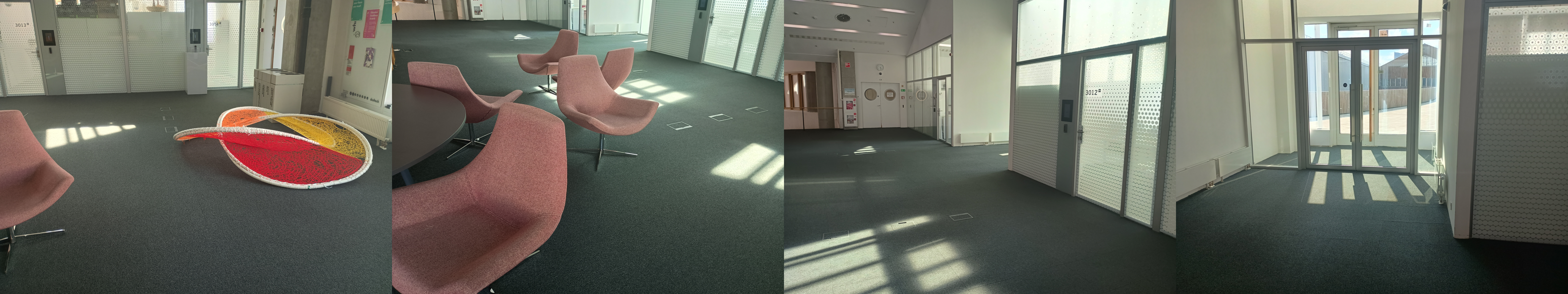}
        \caption{Space \#1}
        \label{fig:space_1}
    \end{subfigure}
    
    \vspace{0.5cm} 
    
    \begin{subfigure}[b]{0.9\linewidth}
        \centering
        \includegraphics[width=0.9\textwidth]{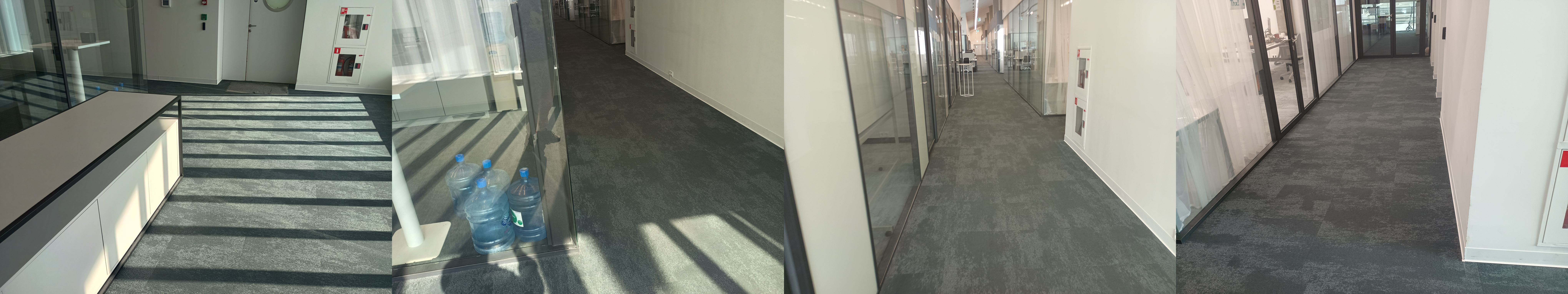}
        \caption{Space \#2}
        \label{fig:space_2}
    \end{subfigure}
    
    \caption{Spaces selected for the evaluation}
    \label{fig:spaces}
\end{figure}

We select two challenging indoor locations at a university campus, referred to as \text{Space 1} and \textit{Space 2} (see Fig. \ref{fig:spaces}). In each of them, an expert path is recorded, and a topological graph is built from the recorded data. The evaluation begins approximately 12-24 hours after recording to ensure the methods' robustness to the lighting changes and minor environment updates.

For each location and each method, we first evaluate the performance of the method in \textit{no perturbation} setting: the state of the environment is similar to the one from the recorded topological graph. The goal of the model is to follow an expert path with the same start and end points. Then, based on this evaluation, a \textit{perturbation} is introduced: a remarkable obstacle, which we call \textit{target obstacle}, is placed in some region that the method tended to cross during the \textit{no perturbation} scenario. In this way, we can evaluate the method's collision avoidance and goal-reaching capabilities when a previously unseen obstacle with a high probability of collision is introduced. We perform two separate perturbations for each location, named \textit{Perturbation 1} and \textit{Perturbation 2}. For each $\textit{no perturbation}$ and $\textit{perturbation}$ setting, we give 3 trials for each method, resulting in 9 runs for each location-method pair.

\begin{figure}[htbp]
    \centering
    \begin{subfigure}[b]{0.9\linewidth}
        \centering
        \includegraphics[width=0.9\textwidth]{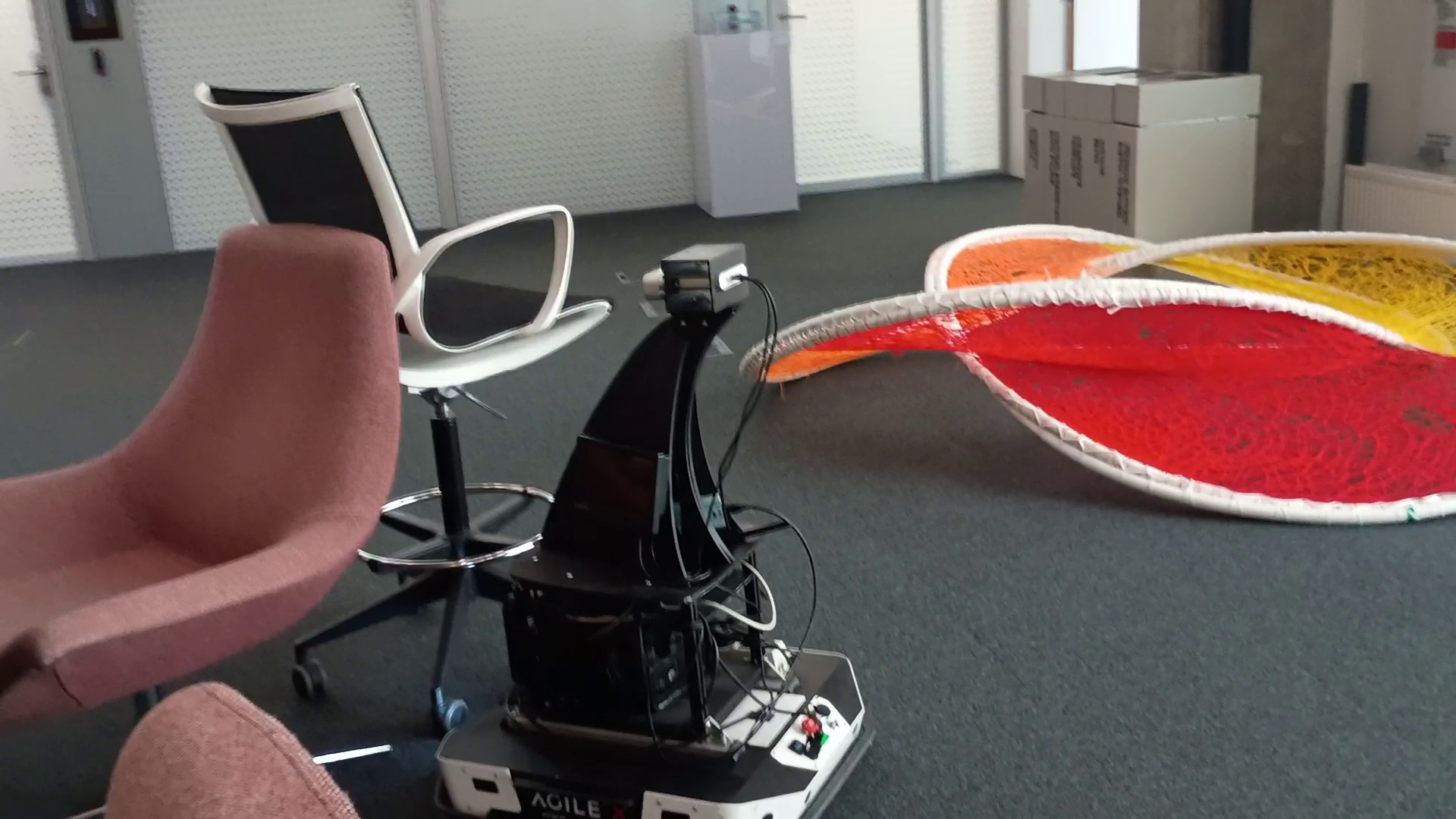}
        \caption{Direct collision}
        \label{fig:dc_example}
    \end{subfigure}
    
    \vspace{0.5cm} 
    
    \begin{subfigure}[b]{0.9\linewidth}
        \centering
        \includegraphics[width=0.9\textwidth]{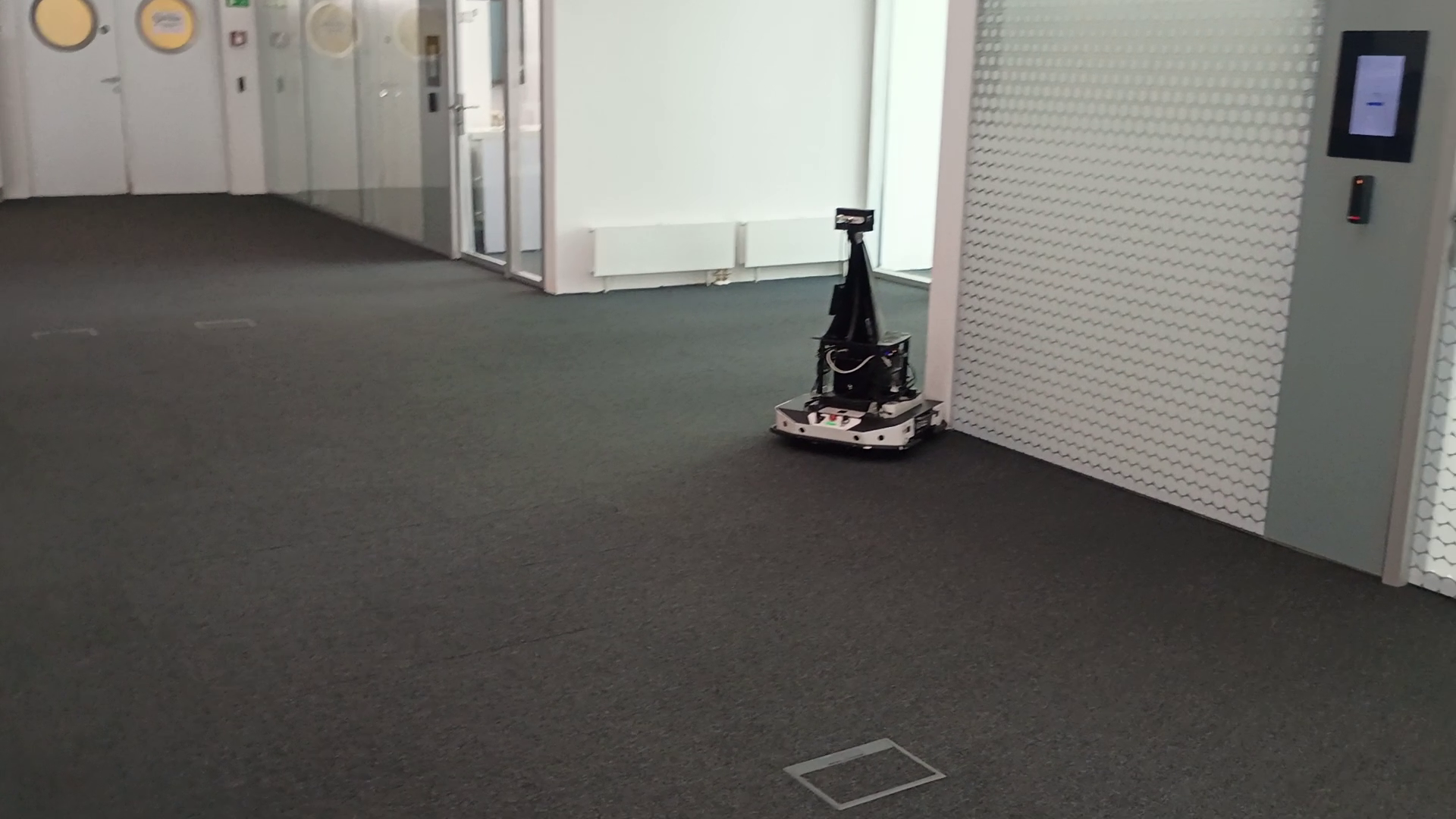}
        \caption{Indirect collision}
        \label{fig:ic_example}
    \end{subfigure}
    
    \caption{Direct and indirect collisions examples}
    \label{fig:dc_ic_example}
\end{figure}

Since existing vision-only approaches have limited temporal and spatial reasoning capabilities, we argue that it is not fair to expect perfect zero-collision performance. Thus, to make an evaluation more fair, we propose to split collision cases into two classes (see Fig. \ref{fig:dc_ic_example} for examples):
\begin{itemize}
    \item \textit{Direct Collisions (DC)}: Collision with the obstacle that is directly visible by the robot's camera at the moment of collision (Fig. \ref{fig:dc_example});
    \item \textit{Indirect Collisions (IC)}: Collision that occurs when the obstacle is not visible (Fig. \ref{fig:ic_example}). For example, the navigation method could select a generally proper maneuver for avoidance, but the robot's base hit after the obstacle left the field of view.
\end{itemize}
Moreover, we additionally count direct collisions with the target obstacle only to evaluate the robustness to the environment perturbations.

\textit{Freezes} are defined as cases when the method explicitly does not produce control inputs or produces idle control inputs, for example, to avoid collision. Freezes in front of the obstacles, unless the robot's base doesn't touch it, are not counted as collisions.

In case of collisions and freezes, human operator's manual intervention is performed to correct the robot's path. If the method selects the completely wrong direction, the robot is considered lost, and the goal is not counted as reached.

Thus, the performance of the methods is evaluated using following metrics:
\begin{itemize}
    \item \underline{A}verage number of \underline{d}irect \underline{c}ollisions per run (\textit{ADC});
    \item \underline{A}verage number of \underline{i}ndirect \underline{c}ollisions per run (\textit{AIC});
    \item \underline{T}arget obstacle \underline{d}irect \underline{c}ollisions rate (\textit{TDCR});
    \item \underline{A}verage number of \underline{f}reezes per run (\textit{AF});
    \item \underline{G}oal \underline{r}eaching \underline{r}ate (\textit{GRR}).
\end{itemize}

\subsection{Baselines}
The end-to-end vision-only navigation models ViNT \cite{shahvint} and NoMaD \cite{sridhar2024nomad} are selected as baseline methods. We use official code and weights releases, both for the topological graph construction and the actual navigation. 

\subsection{Results and Analysis}
The resulting metrics are presented in Tables \ref{tab:results_1} and \ref{tab:results_2}. We show metrics for the spaces \textit{\#1} and \textit{\#2} separately for more fine-grained evaluation. Within the discussed evaluation paradigm, PixelNav significantly outperforms in terms of goal reaching. In the \textit{Space \#1}, a reasonable reaching rate among baselines was obtained only by \textit{NoMaD}, while in \textit{Space \# 2}, both baselines completely failed to select a proper turn at the end of the trajectory. 

\begin{table}[htbp]
\centering
\caption{Results for \textit{Space \#1}}
\label{tab:results_1}
\small
\begin{tblr}{
  hline{1-2,5} = {-}{},
}
\textbf{Method} & \textbf{ADC} $\downarrow$ & \textbf{AIC} $\downarrow$ & \textbf{TDCR} $\downarrow$ & \textbf{AF} $\downarrow$ & \textbf{GRR} $\uparrow$ \\
ViNT            & 0.4                        & 0.0                        & 0.5                         & 0.0                       & 0.2                       \\
NoMaD           & 0.9                        & 0.1                        & 0.5                         &  0.0                      & 0.7                      \\
PixelNav & 0.7                     & 0.7                     & 0.3                      & 0.0                      & 0.9                   
\end{tblr}
\end{table}

\begin{table}[htbp]
\centering
\caption{Results for \textit{Space \#2}}
\label{tab:results_2}
\small
\begin{tblr}{
  hline{1-2,5} = {-}{},
}
\textbf{Method} & \textbf{ADC} $\downarrow$ & \textbf{AIC} $\downarrow$ & \textbf{TDCR} $\downarrow$ & \textbf{AF} $\downarrow$ & \textbf{GRR} $\uparrow$ \\
ViNT            & 0.1                        & 1.3                        & 0.2                         & 0.0                        & 0.0                      \\
NoMaD           & 1.0                        & 0.6                        & 0.5                         & 0.0                       &  0.0                     \\
PixelNav & 0.7                     & 1.1                     & 0.2                      & 0.8                      & 0.8                   
\end{tblr}
\end{table}

We find that the baseline methods tend to ``overfit'' to the expert trajectory; this results in a lower collision rate in general, but significantly limits reaction to the unseen obstacles - these aspects are reflected by the ADC and TDCR values. PixelNav shows higher variance in the behaviour and a more explicit reaction to the unseen obstacles. For the more challenging \textit{Space \# 2}, we observe that the model freezes when it fails to produce proper maneuvers and gets stuck in front of the obstacle due to the lack of traversable area. We argue that this is an acceptable behaviour, since later, some additional rollback policy can be added for such cases. Since PixelNav does not use any form of temporal history of observation, it often touches the obstacles with the edge parts of the base, thus achieving high AIC values. However, the temporal context used in ViNT and NoMaD also does not always help to address this issue.

\begin{figure}[htbp]
    \centering
    \begin{subfigure}[b]{0.9\linewidth}
        \centering
        \includegraphics[width=0.9\textwidth]{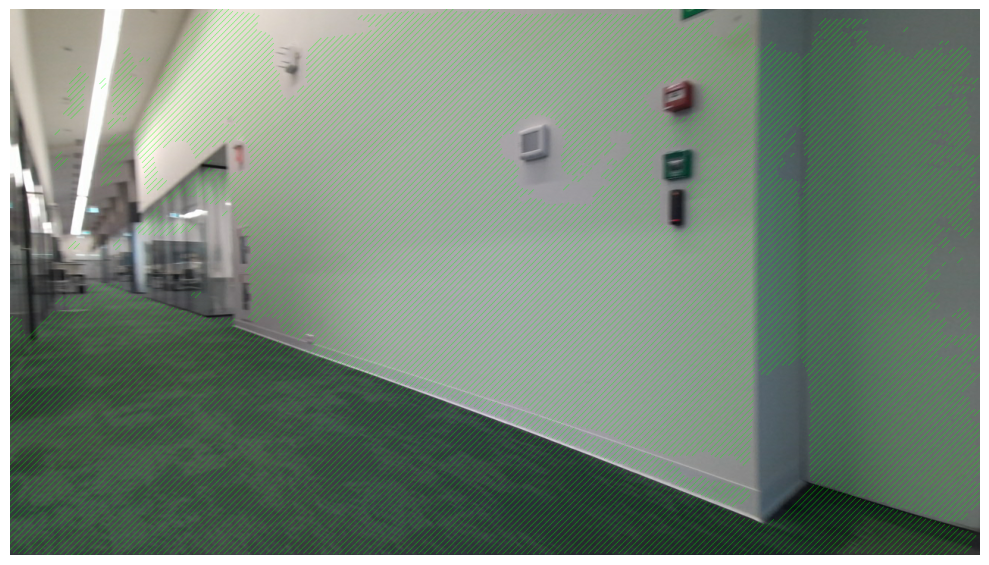}
    \end{subfigure}
    
    \vspace{0.5cm} 
    
    \begin{subfigure}[b]{0.9\linewidth}
        \centering
        \includegraphics[width=0.9\textwidth]{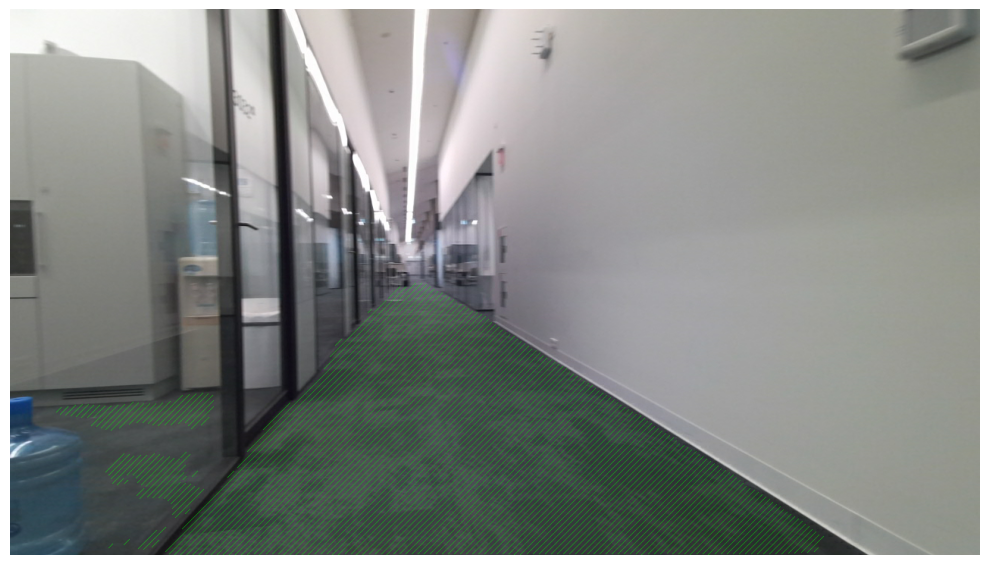}
    \end{subfigure}
    
    \caption{Example of the traversability estimation failures in the evaluation scenes}
    \label{fig:trav_bad_examples}
\end{figure}

In general, we found the following sources of PixelNav's failures:
\begin{itemize}
    \item \textit{Errors in traversability estimation}. We observe that the most challenging cases for the model are when the camera is too close to the walls, especially for the white textureless ones (see Figure \ref{fig:trav_bad_examples}). For some reason, it often confuses white walls with traversable regions.
    \item \textit{Imprecise IPM}. We found that the current implementation of IPM does not give precise mapping and rather acts as a heuristic (this explains the high value of $r^{\text{safe}}$ in Table \ref{tab:model_parameters}). Potentially, it can be improved by a more precise estimation of the camera height.
    \item  \textit{Lack of the temporal context}. Since the model works only with the current observations, it sometimes fails to complete an initially proper maneuver since the obstacle goes away from the field of view. This can especially be seen in examples with the corridor corners and the art object in \textit{Space \#1}. 
\end{itemize}

To summarize, we emphasize that PixelNav performs on par with the modern vision-only baselines, showing a higher goal-reaching rate and robustness to the unseen obstacles. However, due to modular architecture and a model-based controller, we can explicitly identify and fix the bottlenecks.

%% file: sec/5_conclusions.tex
\section{Conclusions}
In this work, a proof-of-concept approach for vision-only navigation that combines VPR, traversability estimation, and MPPI was introduced and evaluated in real-world conditions. It achieves performance comparable to the end-to-end baselines while maintaining a significantly higher level of interpretability, which allows independent improvements of the system components. Future work will focus on those improvements and adding temporal context to the proposed method.

%% file: main.bbl
\begin{thebibliography}{49}
\providecommand{\natexlab}[1]{#1}
\providecommand{\url}[1]{\texttt{#1}}
\expandafter\ifx\csname urlstyle\endcsname\relax
  \providecommand{\doi}[1]{doi: #1}\else
  \providecommand{\doi}{doi: \begingroup \urlstyle{rm}\Url}\fi

\bibitem[Adamkiewicz et~al.(2022)Adamkiewicz, Chen, Caccavale, Gardner, Culbertson, Bohg, and Schwager]{adamkiewicz2022vision}
Michal Adamkiewicz, Timothy Chen, Adam Caccavale, Rachel Gardner, Preston Culbertson, Jeannette Bohg, and Mac Schwager.
\newblock Vision-only robot navigation in a neural radiance world.
\newblock \emph{IEEE Robotics and Automation Letters}, 7\penalty0 (2):\penalty0 4606--4613, 2022.

\bibitem[Akhtyamov et~al.(2025)Akhtyamov, Mdfaa, Ramirez, Bakulin, Devchich, Fatykhov, Mazurov, Zipa, Mohrat, Kolesnik, et~al.]{akhtyamov2025egowalk}
Timur Akhtyamov, Mohamad~Al Mdfaa, Javier~Antonio Ramirez, Sergey Bakulin, German Devchich, Denis Fatykhov, Alexander Mazurov, Kristina Zipa, Malik Mohrat, Pavel Kolesnik, et~al.
\newblock Egowalk: A multimodal dataset for robot navigation in the wild.
\newblock \emph{arXiv preprint arXiv:2505.21282}, 2025.

\bibitem[Arandjelovic et~al.(2016)Arandjelovic, Gronat, Torii, Pajdla, and Sivic]{arandjelovic2016netvlad}
Relja Arandjelovic, Petr Gronat, Akihiko Torii, Tomas Pajdla, and Josef Sivic.
\newblock Netvlad: Cnn architecture for weakly supervised place recognition.
\newblock In \emph{Proceedings of the IEEE conference on computer vision and pattern recognition}, pages 5297--5307, 2016.

\bibitem[Bar et~al.(2025)Bar, Zhou, Tran, Darrell, and LeCun]{bar2025navigation}
Amir Bar, Gaoyue Zhou, Danny Tran, Trevor Darrell, and Yann LeCun.
\newblock Navigation world models.
\newblock In \emph{Proceedings of the Computer Vision and Pattern Recognition Conference}, pages 15791--15801, 2025.

\bibitem[Cai et~al.(2022)Cai, Everett, Fink, and How]{cai2022risk}
Xiaoyi Cai, Michael Everett, Jonathan Fink, and Jonathan~P How.
\newblock Risk-aware off-road navigation via a learned speed distribution map.
\newblock In \emph{2022 IEEE/RSJ International Conference on Intelligent Robots and Systems (IROS)}, pages 2931--2937. IEEE, 2022.

\bibitem[Cai et~al.(2023)Cai, Everett, Sharma, Osteen, and How]{cai2023probabilistic}
Xiaoyi Cai, Michael Everett, Lakshay Sharma, Philip~R Osteen, and Jonathan~P How.
\newblock Probabilistic traversability model for risk-aware motion planning in off-road environments.
\newblock In \emph{2023 IEEE/RSJ International Conference on Intelligent Robots and Systems (IROS)}, pages 11297--11304. IEEE, 2023.

\bibitem[Cai et~al.(2024)Cai, Ancha, Sharma, Osteen, Bucher, Phillips, Wang, Everett, Roy, and How]{cai2024evora}
Xiaoyi Cai, Siddharth Ancha, Lakshay Sharma, Philip~R Osteen, Bernadette Bucher, Stephen Phillips, Jiuguang Wang, Michael Everett, Nicholas Roy, and Jonathan~P How.
\newblock Evora: Deep evidential traversability learning for risk-aware off-road autonomy.
\newblock \emph{IEEE Transactions on Robotics}, 2024.

\bibitem[Chen et~al.(2025)Chen, Shorinwa, Bruno, Swann, Yu, Zeng, Nagami, Dames, and Schwager]{chen2025splat}
Timothy Chen, Ola Shorinwa, Joseph Bruno, Aiden Swann, Javier Yu, Weijia Zeng, Keiko Nagami, Philip Dames, and Mac Schwager.
\newblock Splat-nav: Safe real-time robot navigation in gaussian splatting maps.
\newblock \emph{IEEE Transactions on Robotics}, 2025.

\bibitem[Cheng et~al.(2023)Cheng, Shen, Zhu, Guo, Fang, Liu, Du, and Tao]{cheng2023prescribed}
Zhihao Cheng, Li Shen, Miaoxi Zhu, Jiaxian Guo, Meng Fang, Liu Liu, Bo Du, and Dacheng Tao.
\newblock Prescribed safety performance imitation learning from a single expert dataset.
\newblock \emph{IEEE transactions on pattern analysis and machine intelligence}, 45\penalty0 (10):\penalty0 12236--12249, 2023.

\bibitem[Chiang et~al.(2024)Chiang, Xu, Fu, Jacob, Zhang, Lee, Yu, Schenck, Rendleman, Shah, et~al.]{chiang2024mobility}
Hao-Tien~Lewis Chiang, Zhuo Xu, Zipeng Fu, Mithun~George Jacob, Tingnan Zhang, Tsang-Wei~Edward Lee, Wenhao Yu, Connor Schenck, David Rendleman, Dhruv Shah, et~al.
\newblock Mobility vla: Multimodal instruction navigation with long-context vlms and topological graphs.
\newblock \emph{arXiv preprint arXiv:2407.07775}, 2024.

\bibitem[Cosner et~al.(2022)Cosner, Yue, and Ames]{cosner2022end}
Ryan~K Cosner, Yisong Yue, and Aaron~D Ames.
\newblock End-to-end imitation learning with safety guarantees using control barrier functions.
\newblock In \emph{2022 IEEE 61st Conference on Decision and Control (CDC)}, pages 5316--5322. IEEE, 2022.

\bibitem[DeTone et~al.(2018)DeTone, Malisiewicz, and Rabinovich]{detone2018superpoint}
Daniel DeTone, Tomasz Malisiewicz, and Andrew Rabinovich.
\newblock Superpoint: Self-supervised interest point detection and description.
\newblock In \emph{Proceedings of the IEEE conference on computer vision and pattern recognition workshops}, pages 224--236, 2018.

\bibitem[Dimitropoulos et~al.(2022)Dimitropoulos, Hatzilygeroudis, and Chatzilygeroudis]{dimitropoulos2022brief}
Konstantinos Dimitropoulos, Ioannis Hatzilygeroudis, and Konstantinos Chatzilygeroudis.
\newblock A brief survey of sim2real methods for robot learning.
\newblock In \emph{International Conference on Robotics in Alpe-Adria Danube Region}, pages 133--140. Springer, 2022.

\bibitem[Everett et~al.(2021)Everett, L{\"u}tjens, and How]{everett2021certifiable}
Michael Everett, Bj{\"o}rn L{\"u}tjens, and Jonathan~P How.
\newblock Certifiable robustness to adversarial state uncertainty in deep reinforcement learning.
\newblock \emph{IEEE Transactions on Neural Networks and Learning Systems}, 33\penalty0 (9):\penalty0 4184--4198, 2021.

\bibitem[Fischler and Bolles(1981)]{fischler1981random}
Martin~A Fischler and Robert~C Bolles.
\newblock Random sample consensus: a paradigm for model fitting with applications to image analysis and automated cartography.
\newblock \emph{Communications of the ACM}, 24\penalty0 (6):\penalty0 381--395, 1981.

\bibitem[Gasparino et~al.(2022)Gasparino, Sivakumar, Liu, Velasquez, Higuti, Rogers, Tran, and Chowdhary]{gasparino2022wayfast}
Mateus~V Gasparino, Arun~N Sivakumar, Yixiao Liu, Andres~EB Velasquez, Vitor~AH Higuti, John Rogers, Huy Tran, and Girish Chowdhary.
\newblock Wayfast: Navigation with predictive traversability in the field.
\newblock \emph{IEEE Robotics and Automation Letters}, 7\penalty0 (4):\penalty0 10651--10658, 2022.

\bibitem[Gasparino et~al.(2024)Gasparino, Sivakumar, and Chowdhary]{gasparino2024wayfaster}
Mateus~V Gasparino, Arun~N Sivakumar, and Girish Chowdhary.
\newblock Wayfaster: a self-supervised traversability prediction for increased navigation awareness.
\newblock In \emph{2024 IEEE International Conference on Robotics and Automation (ICRA)}, pages 8486--8492. IEEE, 2024.

\bibitem[Gode et~al.(2024)Gode, Nayak, and Burgard]{gode2024flownav}
Samiran Gode, Abhijeet Nayak, and Wolfram Burgard.
\newblock Flownav: Learning efficient navigation policies via conditional flow matching.
\newblock In \emph{2nd CoRL Workshop on Learning Effective Abstractions for Planning}, 2024.

\bibitem[Hirose et~al.(2023)Hirose, Shah, Sridhar, and Levine]{hirose2023sacson}
Noriaki Hirose, Dhruv Shah, Ajay Sridhar, and Sergey Levine.
\newblock Sacson: Scalable autonomous control for social navigation.
\newblock \emph{IEEE Robotics and Automation Letters}, 9\penalty0 (1):\penalty0 49--56, 2023.

\bibitem[Hirose et~al.(2025)Hirose, Glossop, Sridhar, Mees, and Levine]{hirose2025lelan}
Noriaki Hirose, Catherine Glossop, Ajay Sridhar, Oier Mees, and Sergey Levine.
\newblock Lelan: Learning a language-conditioned navigation policy from in-the-wild video.
\newblock In \emph{Conference on Robot Learning}, pages 666--688. PMLR, 2025.

\bibitem[Jung et~al.(2024)Jung, Lee, Meng, Boots, and Lambert]{jung2024v}
Sanghun Jung, JoonHo Lee, Xiangyun Meng, Byron Boots, and Alexander Lambert.
\newblock V-strong: Visual self-supervised traversability learning for off-road navigation.
\newblock In \emph{2024 IEEE International Conference on Robotics and Automation (ICRA)}, pages 1766--1773. IEEE, 2024.

\bibitem[Karnan et~al.(2022)Karnan, Nair, Xiao, Warnell, Pirk, Toshev, Hart, Biswas, and Stone]{karnan2022socially}
Haresh Karnan, Anirudh Nair, Xuesu Xiao, Garrett Warnell, S{\"o}ren Pirk, Alexander Toshev, Justin Hart, Joydeep Biswas, and Peter Stone.
\newblock Socially compliant navigation dataset (scand): A large-scale dataset of demonstrations for social navigation.
\newblock \emph{IEEE Robotics and Automation Letters}, 7\penalty0 (4):\penalty0 11807--11814, 2022.

\bibitem[Kim et~al.(2024)Kim, Lee, Lee, Mun, Youm, Park, and Hwangbo]{kim2024learning}
Yunho Kim, Jeong~Hyun Lee, Choongin Lee, Juhyeok Mun, Donghoon Youm, Jeongsoo Park, and Jemin Hwangbo.
\newblock Learning semantic traversability with egocentric video and automated annotation strategy.
\newblock \emph{IEEE Robotics and Automation Letters}, 2024.

\bibitem[Kirillov et~al.(2023)Kirillov, Mintun, Ravi, Mao, Rolland, Gustafson, Xiao, Whitehead, Berg, Lo, et~al.]{kirillov2023segment}
Alexander Kirillov, Eric Mintun, Nikhila Ravi, Hanzi Mao, Chloe Rolland, Laura Gustafson, Tete Xiao, Spencer Whitehead, Alexander~C Berg, Wan-Yen Lo, et~al.
\newblock Segment anything.
\newblock In \emph{Proceedings of the IEEE/CVF international conference on computer vision}, pages 4015--4026, 2023.

\bibitem[Kulh{\'a}nek et~al.(2019)Kulh{\'a}nek, Derner, De~Bruin, and Babu{\v{s}}ka]{kulhanek2019vision}
Jon{\'a}{\v{s}} Kulh{\'a}nek, Erik Derner, Tim De~Bruin, and Robert Babu{\v{s}}ka.
\newblock Vision-based navigation using deep reinforcement learning.
\newblock In \emph{2019 european conference on mobile robots (ECMR)}, pages 1--8. IEEE, 2019.

\bibitem[Li et~al.(2022)Li, Xiong, Li, Wu, Zhang, Liu, Bian, and Dou]{li2022interpretable}
Xuhong Li, Haoyi Xiong, Xingjian Li, Xuanyu Wu, Xiao Zhang, Ji Liu, Jiang Bian, and Dejing Dou.
\newblock Interpretable deep learning: Interpretation, interpretability, trustworthiness, and beyond.
\newblock \emph{Knowledge and Information Systems}, 64\penalty0 (12):\penalty0 3197--3234, 2022.

\bibitem[Liu et~al.(2025)Liu, Li, Jiang, Sujay, Yang, Zhang, Abanes, Zhang, and Feng]{liu2025citywalker}
Xinhao Liu, Jintong Li, Yicheng Jiang, Niranjan Sujay, Zhicheng Yang, Juexiao Zhang, John Abanes, Jing Zhang, and Chen Feng.
\newblock Citywalker: Learning embodied urban navigation from web-scale videos.
\newblock In \emph{Proceedings of the Computer Vision and Pattern Recognition Conference}, pages 6875--6885, 2025.

\bibitem[Macenski et~al.(2022)Macenski, Foote, Gerkey, Lalancette, and Woodall]{doi:10.1126/scirobotics.abm6074}
Steven Macenski, Tully Foote, Brian Gerkey, Chris Lalancette, and William Woodall.
\newblock Robot operating system 2: Design, architecture, and uses in the wild.
\newblock \emph{Science Robotics}, 7\penalty0 (66):\penalty0 eabm6074, 2022.

\bibitem[McCarthy and Bames(2004)]{mccarthy2004performance}
Chris McCarthy and Nick Bames.
\newblock Performance of optical flow techniques for indoor navigation with a mobile robot.
\newblock In \emph{IEEE International Conference on Robotics and Automation, 2004. Proceedings. ICRA'04. 2004}, pages 5093--5098. IEEE, 2004.

\bibitem[Nguyen et~al.(2023)Nguyen, Nazeri, Payandeh, Datar, and Xiao]{nguyen2023toward}
Duc~M Nguyen, Mohammad Nazeri, Amirreza Payandeh, Aniket Datar, and Xuesu Xiao.
\newblock Toward human-like social robot navigation: A large-scale, multi-modal, social human navigation dataset.
\newblock In \emph{2023 IEEE/RSJ International Conference on Intelligent Robots and Systems (IROS)}, pages 7442--7447. IEEE, 2023.

\bibitem[Ohnishi and Imiya(2008)]{ohnishi2008visual}
Naoya Ohnishi and Atsushi Imiya.
\newblock Visual navigation of mobile robot using optical flow and visual potential field.
\newblock In \emph{International Workshop on Robot Vision}, pages 412--426. Springer, 2008.

\bibitem[Oquab et~al.(2024)Oquab, Darcet, Moutakanni, Vo, Szafraniec, Khalidov, Fernandez, Haziza, Massa, El-Nouby, et~al.]{oquab2024dinov2}
Maxime Oquab, Timoth{\'e}e Darcet, Th{\'e}o Moutakanni, Huy Vo, Marc Szafraniec, Vasil Khalidov, Pierre Fernandez, Daniel Haziza, Francisco Massa, Alaaeldin El-Nouby, et~al.
\newblock Dinov2: Learning robust visual features without supervision.
\newblock \emph{Transactions on Machine Learning Research Journal}, pages 1--31, 2024.

\bibitem[Phan et~al.(2025)Phan, Nguyen, Au, Phan, Duong, and Le]{phan2025visionbasedperceptionautonomousvehicles}
Van-Hoang-Anh Phan, Chi-Tam Nguyen, Doan-Trung Au, Thanh-Danh Phan, Minh-Thien Duong, and My-Ha Le.
\newblock Vision-based perception for autonomous vehicles in obstacle avoidance scenarios, 2025.

\bibitem[Roth et~al.(2024)Roth, Nubert, Yang, Mittal, and Hutter]{roth2024viplanner}
Pascal Roth, Julian Nubert, Fan Yang, Mayank Mittal, and Marco Hutter.
\newblock Viplanner: Visual semantic imperative learning for local navigation.
\newblock In \emph{2024 IEEE International Conference on Robotics and Automation (ICRA)}, pages 5243--5249. IEEE, 2024.

\bibitem[Sarlin et~al.(2020)Sarlin, DeTone, Malisiewicz, and Rabinovich]{sarlin2020superglue}
Paul-Edouard Sarlin, Daniel DeTone, Tomasz Malisiewicz, and Andrew Rabinovich.
\newblock Superglue: Learning feature matching with graph neural networks.
\newblock In \emph{Proceedings of the IEEE/CVF conference on computer vision and pattern recognition}, pages 4938--4947, 2020.

\bibitem[Savinov et~al.(2018)Savinov, Dosovitskiy, and Koltun]{savinov2018semi}
Nikolay Savinov, Alexey Dosovitskiy, and Vladlen Koltun.
\newblock Semi-parametric topological memory for navigation.
\newblock In \emph{International Conference on Learning Representations}, 2018.

\bibitem[Shah and Levine(2022)]{shah2022viking}
Dhruv Shah and Sergey Levine.
\newblock Viking: Vision-based kilometer-scale navigation with geographic hints.
\newblock \emph{arXiv preprint arXiv:2202.11271}, 2022.

\bibitem[Shah et~al.(2021{\natexlab{a}})Shah, Eysenbach, Kahn, Rhinehart, and Levine]{shah2021rapid}
Dhruv Shah, Benjamin Eysenbach, Gregory Kahn, Nicholas Rhinehart, and Sergey Levine.
\newblock Rapid exploration for open-world navigation with latent goal models.
\newblock \emph{arXiv preprint arXiv:2104.05859}, 2021{\natexlab{a}}.

\bibitem[Shah et~al.(2021{\natexlab{b}})Shah, Eysenbach, Kahn, Rhinehart, and Levine]{shah2021ving}
Dhruv Shah, Benjamin Eysenbach, Gregory Kahn, Nicholas Rhinehart, and Sergey Levine.
\newblock Ving: Learning open-world navigation with visual goals.
\newblock In \emph{2021 IEEE International Conference on Robotics and Automation (ICRA)}, pages 13215--13222. IEEE, 2021{\natexlab{b}}.

\bibitem[Shah et~al.(2022)Shah, Sridhar, Bhorkar, Hirose, and Levine]{shah2022gnm}
Dhruv Shah, Ajay Sridhar, Arjun Bhorkar, Noriaki Hirose, and Sergey Levine.
\newblock Gnm: A general navigation model to drive any robot.
\newblock \emph{arXiv preprint arXiv:2210.03370}, 2022.

\bibitem[Shah et~al.(2023)Shah, Sridhar, Dashora, Stachowicz, Black, Hirose, and Levine]{shahvint}
Dhruv Shah, Ajay Sridhar, Nitish Dashora, Kyle Stachowicz, Kevin Black, Noriaki Hirose, and Sergey Levine.
\newblock Vint: A foundation model for visual navigation.
\newblock In \emph{7th Annual Conference on Robot Learning}, 2023.

\bibitem[Sridhar et~al.(2024)Sridhar, Shah, Glossop, and Levine]{sridhar2024nomad}
Ajay Sridhar, Dhruv Shah, Catherine Glossop, and Sergey Levine.
\newblock Nomad: Goal masked diffusion policies for navigation and exploration.
\newblock In \emph{2024 IEEE International Conference on Robotics and Automation (ICRA)}, pages 63--70. IEEE, 2024.

\bibitem[Suzuki et~al.(1985)]{suzuki1985topological}
Satoshi Suzuki et~al.
\newblock Topological structural analysis of digitized binary images by border following.
\newblock \emph{Computer vision, graphics, and image processing}, 30\penalty0 (1):\penalty0 32--46, 1985.

\bibitem[Teed and Deng(2022)]{dpvo2022}
Zachary Teed and Jia Deng.
\newblock Deep patch visual odometry.
\newblock In \emph{European Conference on Computer Vision}, pages 460--477. Springer, 2022.

\bibitem[Wang et~al.(2023)Wang, Cao, Yu, Wang, Fu, Yang, Zhou, and Geiger]{anyloc2023}
Qi Wang, Zixin Cao, Yifan Yu, Zhichao Wang, Chang Fu, Xin Yang, Hang Zhou, and Andreas Geiger.
\newblock Anyloc: A foundation model for long-term visual place recognition.
\newblock \emph{arXiv preprint arXiv:2307.16849}, 2023.

\bibitem[Williams et~al.(2017)Williams, Wagener, Goldfain, Drews, Rehg, Boots, and Theodorou]{williams2017information}
Grady Williams, Nolan Wagener, Brian Goldfain, Paul Drews, James~M Rehg, Byron Boots, and Evangelos~A Theodorou.
\newblock Information theoretic mpc for model-based reinforcement learning.
\newblock In \emph{2017 IEEE international conference on robotics and automation (ICRA)}, pages 1714--1721. IEEE, 2017.

\bibitem[Xie et~al.(2021)Xie, Wang, Yu, Anandkumar, Alvarez, and Luo]{xie2021segformer}
Enze Xie, Wenhai Wang, Zhiding Yu, Anima Anandkumar, Jose~M Alvarez, and Ping Luo.
\newblock Segformer: Simple and efficient design for semantic segmentation with transformers.
\newblock \emph{Advances in neural information processing systems}, 34:\penalty0 12077--12090, 2021.

\bibitem[Zeng et~al.(2020)Zeng, Wang, and Ge]{zeng2020survey}
Fanyu Zeng, Chen Wang, and Shuzhi~Sam Ge.
\newblock A survey on visual navigation for artificial agents with deep reinforcement learning.
\newblock \emph{IEEE Access}, 8:\penalty0 135426--135442, 2020.

\bibitem[Zhu et~al.(2017)Zhu, Mottaghi, Kolve, Lim, Gupta, Fei-Fei, and Farhadi]{zhu2017target}
Yuke Zhu, Roozbeh Mottaghi, Eric Kolve, Joseph~J Lim, Abhinav Gupta, Li Fei-Fei, and Ali Farhadi.
\newblock Target-driven visual navigation in indoor scenes using deep reinforcement learning.
\newblock In \emph{2017 IEEE international conference on robotics and automation (ICRA)}, pages 3357--3364. IEEE, 2017.

\end{thebibliography}
